\def\year{2022}\relax
\documentclass[letterpaper]{article} 
\usepackage{aaai22}  
\usepackage{times}  
\usepackage{helvet}  
\usepackage{courier}  
\usepackage[hyphens]{url}  
\usepackage{graphicx} 
\usepackage{natbib}  
\usepackage{caption} 
\DeclareCaptionStyle{ruled}{labelfont=normalfont,labelsep=colon,strut=off} 
\usepackage{color}
\usepackage{array}
\usepackage{makecell}
\usepackage{booktabs}
\usepackage{tabularx}
\usepackage{amsfonts}
\usepackage[ruled, linesnumbered]{algorithm2e}

\usepackage{tikz}

\usepackage{amsmath}
\usepackage{amsthm}
\usepackage{graphicx}
\usepackage{soul}
\usepackage{cite}

\usepackage{enumitem}
\setlist{nosep}
\usepackage{wrapfig}
\usepackage{multirow}

\DeclareMathOperator*{\argmin}{\arg\!\min}

\usepackage{color}

\newtheorem{definition}{Definition}
\newtheorem{theorem}{Theorem}

\newtheorem{lemma}{Lemma}

\newtheorem{formulation}{Problem Formulation}

\theoremstyle{definition}

\newcommand{\X}{\mathbf{X}}
\newcommand{\x}{\mathbf{x}}

\newcommand{\Z}{\mathbf{Z}}
\newcommand{\z}{\mathbf{z}}
\usepackage{dsfont}

\usepackage{mathtools}

\title{Achieving Long-Term Fairness in Sequential Decision Making}
\author{
    Yaowei Hu,
    Lu Zhang 
}
\affiliations{
    University of Arkansas \\
    \{yaoweihu, lz006\}@uark.edu
}

\usepackage{bibentry}

\begin{document}

\maketitle

\begin{abstract}
In this paper, we propose a framework for achieving long-term fair sequential decision making. By conducting both the hard and soft interventions, we propose to take path-specific effects on the time-lagged causal graph as a quantitative tool for measuring long-term fairness. The problem of fair sequential decision making is then formulated as a constrained optimization problem with the utility as the objective and the long-term and short-term fairness as constraints. We show that such an optimization problem can be converted to a performative risk optimization. Finally, repeated risk minimization (RRM) is used for model training, and the convergence of RRM is theoretically analyzed. The empirical evaluation shows the effectiveness of the proposed algorithm on synthetic and semi-synthetic temporal datasets.
\end{abstract}

\section{Introduction}
Fair machine learning has received increasing attention in the past years, especially in decision making tasks such as hiring \cite{Caton2020FairnessIM}, college admissions \cite{Zhang2017AchievingNI} and bank loans \cite{johnson2016impartial}. 
Many algorithms for achieving fair decision making have been proposed based on various fairness notions (e.g. demographic parity \cite{zemel2013learning}, equalized odds \cite{Hardt2016EqualityOO} and counterfactual fairness \cite{Kusner2017CounterfactualF}).
At present, the majority of studies on fair machine learning focus on the static or one-shot classification setting. 
However, in practice, decision making systems are usually operating in a dynamic manner such that the classifier makes sequential decisions over a period of time. In many situations, each decision made by the classifier may change the underlying data population and further affect subsequent decisions. For example, suppose a person applies to a bank for a loan and the bank estimates the risk of default according to his/her credit score. Then, the bank's decision on the loan application (e.g., whether to grant the loan and the interest rate assigned) may in turn affect the default risk and change the person's credit score (e.g., the credit score will decrease if the loan is granted but he/she defaults on the loan) which will affect his/her next loan application. If the bank's decision leads to a long-term decrease in the credit score, then it imposes a negative long-term effect on future decisions for this person. Therefore, fair decision making should concern not only the fairness of a single decision but more importantly, whether a decision model can impose fair long-term effects on different groups. This notion of fairness is referred to as long-term fairness in recent studies \cite{liu2018delayed,hu2018short,ge2021towards}.

The challenge of achieving long-term fairness comes in two folds. Firstly, different from static settings, decisions made by models may change users' behaviors, and/or affect their status such as reputation, qualification, etc., and impact subsequent decisions via feedback loops. Without knowing how the population would be reshaped by decisions, enforcing any fairness constraint may create negative feedback loops and eventually harm fairness in the long run. Recent research has demonstrated that existing fairness criteria cannot guarantee fairness and sometimes undermine fairness even if only one time step is taken into consideration \cite{liu2018delayed,kannan2019downstream,d2020fairness, creager2020causal}. Secondly, due to the feedback loops, the deployment of the decision model will cause changes in the data distribution that is originally used for training. This can be viewed as a distribution shift problem as the distribution of the training data (i.e., distribution before the model deployment) is different from the distribution of the test data (i.e., distribution after the model deployment). Ignoring the distribution shift will critically affect the achievement of long-term fairness, as long-term fairness is affected by all decisions made by the model along the time.

In this paper, we propose a framework for achieving long-term fair sequential decision making by addressing both above challenges. We model the dynamics of the decision-making process by employing Pearl's Structural Causal Model (SCM) \cite{pearl2009causality}, in which the relations among user features and decisions and how those decisions affect the data distribution can be encoded in a probabilistic graphical model. Specifically, we leverage the time-lagged causal graph \cite{runge2019detecting} to describe the causal relations over time, and adopt the soft intervention \cite{correa2020calculus} for modeling the model deployment and inferring its impacts on the underlying population. Then, we measure long-term fairness as the path-specific effect on the time-lagged causal graph under both the hard intervention on the sensitive attribute and the soft intervention on the predicted decisions. A constrained optimization problem is formulated to strike a trade-off between long-term fairness and model utility, as well as certain short-term fairness requirement that may be stipulated by law or regulations. 
On the other hand, we show that the constrained optimization problem can be converted to a performative risk optimization problem \cite{perdomo2020performative}. Then, we employ the repeated risk minimization (RRM) training technique \cite{perdomo2020performative} for dealing with the distribution shift problem. The performative optimality and stability
properties of the proposed method are theoretically and empirically evaluated which shows its effectiveness.

To the best of our knowledge, this paper is the first to propose a causality-based long-fairness notion. The proposed learning framework is general such that it could incorporate different combinations of surrogate functions, utility loss functions, as well as causal paths regarding long-term fairness used to fit different applications. The experiment results show that the proposed method can achieve long-term fairness for multiple time steps, while the fairness performance deteriorates with time if no fairness constraint or static fairness constraints are used. 

{\bf\noindent Related Work.}
Fair machine learning research in past years has been focused on static settings with one-shot decisions being made. To extend fair machine learning to dynamic
settings, some efforts have been devoted to a compound decision-making process called pipeline \cite{bower2017fair,dwork2018fairness}. In pipelines, individuals may drop out at any stage and classification in subsequent stages depends on the remaining cohort of individuals. In a more general setting, decisions made in the past will affect the underlying population, and then affect future decisions \cite{zhang2020fairness}. In this setting, a number of studies have demonstrated the inadequacy of static fairness approaches in various scenarios, including credit lending \cite{liu2018delayed}, college admission \cite{d2020fairness}, labor market \cite{hu2018short}. \citet{creager2020causal} proposes to use causal directed acyclic graphs (DAGs) as a unifying framework to study fairness in dynamical systems, but does not propose an approach to achieve long-term fairness. On the other hand, some works \cite{jabbari2017fairness,zhang2020fair,wen2021algorithms} study long-term fairness in the context of reinforcement learning. The most relevant work to this paper is \cite{hu2020fair} on fair multiple decision making, which also applies SCM and leverages soft interventions to model the deployment of decision models. However, \cite{hu2020fair} is still focused on the static fairness of each decision model separately other than the long-term fairness.

\section{Preliminaries}
Throughout this paper, variables and their values are denoted by uppercase and lowercase letters respectively, i.e., $X$ and $x$. The sets of variables and their values are denoted by bold letters, i.e., $\mathbf{X}$ and $\mathbf{x}$.

\subsection{Structural Causal Model}
Our work is based on Pearl's structural causal models \cite{pearl2009causality} which describe the causal mechanisms of a system by a set of structural equations, i.e., $x = f_X(\mathbf{pa}_X, \mathbf{u}_X)$ for each $X$, where $\mathbf{pa}_{X}$ is a realization of a subset of endogenous variables, and $\mathbf{u}_X$ is a realization of a set of exogenous variables. Each causal model $\mathcal{M}$ is associated with a causal model graph $\mathcal{G = \langle \mathbf{V}, \mathbf{E} \rangle}$ where $\mathbf{V}$ is a set of nodes and $\mathbf{E}$ is a set of directed edges for representing the direct causal relations. This paper assumes the Markovian model in which all exogenous variables are mutually independent.

Quantitatively measuring causal effects is facilitated with the (hard) intervention \cite{pearl2009causality} which forces some variables to take certain values. Formally, the intervention that sets the value of $X$ to $x$ is denoted by $do(x)$. In a SCM, intervention $do(x)$ is defined as the substitution of equation $x = f_X(\mathbf{pa}_X, \mathbf{u}_X)$ with constant $x$. An intervention on a variable affects its descendants via causal paths. For an observed variable $Y$ affected by $X$, its variant under intervention $do(x)$ is denoted by $Y(x)$. The distribution of $Y(x)$, also referred to as the post-intervention distribution of $Y$, is denoted by $P(Y(x))$. The soft intervention \cite{pearl2009causality} extends the hard intervention such that it forces variable $X$ to take functional relationship $g(\mathbf{z})$ in responding to some other variables $\mathbf{Z}$, which is denoted by $\sigma$ in \cite{correa2020calculus}. The soft intervention substitutes equation $x = f_X(\mathbf{pa}_X, \mathbf{u}_X)$ with a new function $x = g(\mathbf{z})$. After performing the soft intervention, $X$ will be associated with a new distribution determined by function $g$. In this paper, the function $g$ is parameterized by $\theta$, and we denote the new distribution by $P_{\theta}(x|\z)$. We also denote the distribution of $Y$ after performing the soft intervention by $P(Y(\theta))$.

\subsection{Fairness-aware Classification}
The classification problem is to learn a functional mapping $f: \mathcal{X} \mapsto \mathcal{Y}$ from the labeled training data $\{(\mathbf{x}_i, y_i)\}_{i=1}^n$ where $\mathbf{x}_i \in \mathcal{X}$, $y_i \in \mathcal{Y}$ and  $\mathcal{Y} = \{-1, 1\}$ , by minimizing the 0-1 loss function $\mathbb{E}_{\mathbf{X}, Y}[\mathds{1}[f(\mathbf{X}) \ne Y]]$ where $\mathds{1}[\cdot]$ is an indicator function. In general, $f$ is made up of another function $h$ set up in the real number domain, i.e., $h: \mathcal{X} \mapsto \mathbb{R}$ and $\mathds{1}[f(\mathbf{X}) \ne Y] = \mathds{1}[Y h(\mathbf{X}) \ge 0]$. 
Since directly minimizing the indicator is intractable, one can replace it with a smooth and differentiable surrogate function $\phi$. Then, the loss function can can be reformulated as $\mathbb{E}_{\mathbf{X}, Y}[\phi(Y  h(\mathbf{X}))]$.
Similarly, one can also formulate fairness constraints as smoothed expressions using surrogate functions. As a result, fair classification problems can be formulated as constrained optimization problems \cite{wu2019convexity,hu2020fair}. We follow the notations used in \cite{wu2019convexity,hu2020fair} in our formulations.

\section{Formulating Long-term Fairness}

We start by formally formulating the long-term fairness in sequential decision making. Assume we have access to a temporal dataset $\mathcal{D}$ = \{$(S, \mathbf{X}^t, Y^t)\}_{t=1}^{l}$ where $S$ is a time-invariant protected attribute, $\mathbf{X}^t$ is a set of time-dependent unprotected attributes and $Y^t$ is a time-dependent class label. Note that this setting can be viewed as observing the data of a set of individuals at all time steps, or a more general situation where a population is subject to the decision cycles and the data is sampled at each time step. For ease of discussion, we assume both class label and protected attribute are binary variables, i.e., $S=\{s^+, s^-\}$ with $s^+$ denoting the unprotected group and $s^-$ denoting the protected group, and $Y=\{1, -1\}$ with $1$ denoting the positive decision and $-1$ denoting the negative decision, but proposed concepts could be extended to multiple protected attributes and multiple/continuous labels situations. 
A predictive decision model $h_{\theta}(\cdot)$ parameterized by $\theta$ is trained on $\mathcal{D}$. Then, it is deployed to make predicted decisions $\hat{Y}^t$ from $(S,\X^{t})$ repeatedly at each time, i.e., $\hat{Y}^t=1$ if $h_{\theta}(s,\x^{t})\geq 0$ and $\hat{Y}^t=-1$ otherwise, forming a sequential decision making process. Such sequential decision making process is common in practice. For example, a bank repeatedly makes lending decisions based on applicants' profile such as credit score, income, etc., and a predictive policing algorithm repeatedly makes decision about where to send police for patrolling based on the crime discovered in the neighborhood. The ultimate goal of long-term fair machine learning is to ensure that the model $h_{\theta}(\cdot)$ is fair in a long-term stage denoted by $t$*. In this paper, we assume there is sufficient historical training data such that $l \ge t^*$.

\subsection{Causality-based Long-term Fairness}
\begin{figure}[t]\small
	\centering
	\begin{tikzpicture}[scale=1.1]
	\definecolor{blue}{rgb}{0.0, 0.0, 0.0}
	\definecolor{red}{rgb}{0.8, 0.0, 0.0}
	\definecolor{green}{rgb}{0.0, 0.5, 0.0}
	
	\tikzstyle{vertex} = [very thick, circle, draw, inner sep=0pt, minimum size = 8mm]
	\tikzstyle{empty} = [very thick, circle, inner sep=0pt, minimum size = 8mm]
	\tikzstyle{rect} = [very thick, rectangle, draw, inner sep=0pt, minimum size = 6mm]
	\tikzstyle{remp} = [very thick, rectangle, inner sep=0pt, minimum size = 3mm]
	\tikzstyle{edge} = [ultra thick, -stealth, blue]
	
	\node[rect] at (0,3.4) (S) {$S$};
	
	\node[vertex] at (0.0,2.2) (X0) {$\mathbf{X}^1$};
	\node[vertex] at (1.3,2.2) (X1) {$\mathbf{X}^2$};
	\node[vertex] at (2.6,2.2) (X2) {$\mathbf{X}^3$};
	\node[empty]  at (3.6,2.2) (X3) {$...$};
	\node[vertex] at (4.6,2.2) (X*) {$\mathbf{X}^{t^*}$};
	
	\node[vertex] at (0.0,1) (Y0) {$Y^1$};
	\node[vertex] at (1.3,1) (Y1) {$Y^2$};
	\node[vertex] at (2.6,1) (Y2) {$Y^3$};
	\node[empty]  at (3.6,1) (Y3) {$...$};
	\node[vertex] at (4.6,1) (Y*) {$Y^{t^*}$};
	
	\node[rect] at (0.0, 0.0) (G0) {$h_{\theta}$};
	\node[rect] at (1.3, 0.0) (G1) {$h_{\theta}$};
	\node[rect] at (2.6, 0.0) (G2) {$h_{\theta}$};
	\node[remp] at (3.6, 0.0) (G3) {$...$};
	\node[rect] at (4.6, 0.0) (G*) {$h_{\theta}$};
	
    \draw[edge, red] (S) to (X0);
    \draw[edge, green] (S) .. controls (-0.6, 2.8) and (-0.5, 1.6) .. (Y0);
    \draw[edge, green] (S) to (Y1);
    \draw[edge, green] (S) .. controls (2.0, 2.6) and (2.3, 1.5).. (Y2);
    \draw[edge, green] (S) .. controls (5.2, 3.0) and (5.4, 2.8).. (Y*);
    
    \draw[edge, red] (X0) to (X1);
    \draw[edge, red] (X1) to (X2);
    \draw[edge, red] (X2) to (X3);
    \draw[edge, red] (X3) to (X*);
    
    \draw[edge, red] (X0) to (Y0);
    \draw[edge, red] (X1) to (Y1);
    \draw[edge, red] (X2) to (Y2);
    \draw[edge, red] (X*) to (Y*);
    
    \draw[edge, red] (Y0) to (X1);
    \draw[edge, red] (Y1) to (X2);
    \draw[edge, red] (Y2) to (3.2, 1.6);
    \draw[edge, red] (4.0, 1.6) to (X*);
    
    \draw[edge, blue] (G0) to (Y0);
    \draw[edge, blue] (G1) to (Y1);
    \draw[edge, blue] (G2) to (Y2);
    \draw[edge, blue] (G*) to (Y*);
    
    \node[vertex] at (0.0,2.2) (X0) {$\mathbf{X}^1$};
	\node[vertex] at (1.3,2.2) (X1) {$\mathbf{X}^2$};
	\node[vertex] at (2.6,2.2) (X2) {$\mathbf{X}^3$};
	\node[empty]  at (3.6,2.2) (X3) {$...$};
    
    
	\end{tikzpicture}
	\caption{A time-lagged causal graph for sequential decision making. Long-term fairness is captured by paths in red, and short-term fairness is captured by paths in green.}
	\label{fig:example2}
\end{figure}
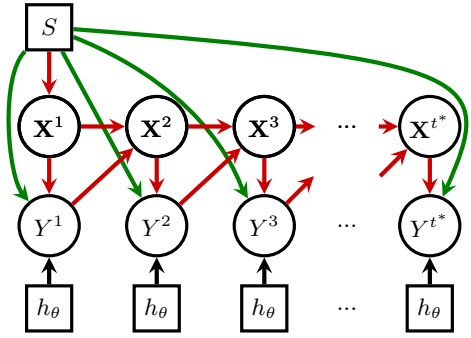

We develop the long-term fairness notion by leveraging Pearl's SCM. First, we assume a time-lagged causal graph $\mathcal{G}$ for describing the causal relationship among variables over time. In recent years, structure learning algorithms have been proposed for constructing time-lagged causal graphs from data, including both constrained-based approaches \cite{runge2018causal,runge2018conditional,runge2020discovering} and continuous optimization-based approaches \cite{pamfil2020dynotears,lowe2020amortized} which can be leveraged to learn the time-lagged causal graph from data. Figure \ref{fig:example2} shows a typical example of the time-lagged causal graph in our settings: 
the edge from $S$ to $\mathbf{X}^0$ represents the bias in the distribution of $\mathbf{X}$ due to historical reasons; the edges from $S$ and $\mathbf{X}^t$ to $Y^t$ represent that $S$ and $\mathbf{X}^t$ are used as the input to compute $\hat{Y}^t$; and the edges from $\mathbf{X}^t$ and $Y^t$ to $\mathbf{X}^{t+1}$ represent how the distribution of $\mathbf{X}$ would be reshaped via feedback after each decision.

Next, we formulate long-term fairness as path-specific effects that are
transmitted in the time-lagged causal graph along certain paths. The path-specific effects reflect how the intervention affects each variable on the path in a topological order and hence are appropriate for capturing dynamics in sequential decision making. Similar to the indirect discrimination in static fair machine learning \cite{zhang2017causal,nabi2018fair}, we can also justify the use of the path-specific effect by the need to distinguish discriminatory effects from explainable effects. We consider discriminatory effects as those which are due to biased decisions made by the decision making system in the past and will continue to influence future decisions. Correspondingly, we consider explainable effects as those which are attributed to external factors and cannot be eliminated within the decision making system. To this end, we categorize unprotected attributes $\X$ into two disjoint subsets: irrelevant attributes $\X_{i}$ and relevant attributes $\X_{r}$. We define irrelevant attributes as those which are justifiable in decision making, and meanwhile evolved autonomously or/and altered by external factors only. We define the rest of attributes as relevant attributes, which could be unjustifiable in decision making or reshaped by the decision over time. Then, we define long-term fairness as the causal effect where the influence of the hard intervention on $S$ is transmitted in the causal graph by passing through relevant attributes only. Note that the influence of the soft intervention on $Y$ is still transmitted through all causal paths.

Finally, we propose to adopt soft interventions as a key technique for modeling decision model deployment and inferring its impacts on the underlying population. We treat the deployment of the decision model at each time step as to perform a soft intervention on the decision variable. 
More specifically, we force the structural equation associated with $Y^{t}$ in the causal model to be replaced by the decision model $h_{\theta}(\cdot)$ that outputs $\hat{Y}^{t}$. Thus, the change to underlying population could be inferred as the post-intervention distribution after performing the soft intervention.
Meanwhile, to quantify fairness as causal effects of the protected attribute on the decision, we perform hard intervention on the protected attribute in order to answer the ``what if'' question, i.e., ``what would the decision be if we intervene the gender of applications to female?'' As a result, we perform both hard intervention and soft intervention simultaneously for measuring long-term fairness as causal effects.

Symbolically, denote by $\pi$ the set of causal paths from $S$ to $\hat{Y}^{t^{*}}$ through relevant attributes $\mathbf{X}_{r}^{1},\cdots,\mathbf{X}_{r}^{t^{*}}$ and $\hat{Y}^{1},\cdots,\hat{Y}^{t^{*}-1}$ but not through irrelevant attributes $\mathbf{X}_{i}^{0},\cdots,\mathbf{X}_{i}^{t^{*}}$. 
Meanwhile, as we conduct path-specific hard intervention on $S$ and soft interventions on $Y$ to deploy decision model $h_{\theta}(\cdot)$, we denote the post-intervention distribution of $\hat{Y}^{t}$ by $\hat{Y}^{t}(s_{\pi},\theta)$ which explicitly shows that the soft intervention depends on parameters $\theta$. Then, we can readily propose the quantitative notion for long-term fairness.

\begin{definition}[Long-term Fairness] 
The long-term fairness of a decision model $h_{\theta}(\cdot)$ is measured by $P(\hat{Y}^{t*}(s^+_{\pi}, \theta)) - P(\hat{Y}^{t*}(s^-_{\pi}, \theta))$ where $\pi$ is a set of paths from $S$ to $\hat{Y}^{t*}$ passing through $\textbf{X}_{r}^{1}$, $\hat{Y}^{1}$, $\cdots$, $\textbf{X}_{r}^{t*-1}$, $\hat{Y}^{t*-1}$, $\textbf{X}_{r}^{t*}$, $s_{\pi}$ represents the path-specific hard intervention and $\theta$ represents the soft intervention through all paths.
\end{definition}

\subsection{Loss Function and Short-term Fairness}
In addition to long-term fairness, a desired fair decision model should also satisfy two other requirements. Firstly, it is a natural desire for a predictive decision model to maximize the institution utility, e.g., the loan granting model of a bank certainly wants to maximize the expected return from loans. 
Secondly, the decision model should also satisfy certain short-term fairness requirement at each time step to enforce local equality, which may be stipulated by law or regulations. For example, the Equal Credit Opportunity Act, 1974, prohibits lending decisions from being influenced by race, age, religion, etc. Similar to the direct discrimination in static fair machine learning, we consider a subset of relevant attributes $\tilde{\X}_{r}\subset \X_{r}$ which are unprotected but cannot be justifiably used in the decision making either directly or indirectly, referred to as the redlining attributes \cite{zhang2017causal}. Then, we measure the short-term fairness by the causal effect of $S$ on $\hat{Y}^t$ along paths that pass through $\tilde{\X}_{r}$, i.e., $S\rightarrow \tilde{\X}_{r} \rightarrow \hat{Y}^{t}$, as well as the direct edge $S \to \hat{Y}^t$ at each time step $t$. 

We note that trade-off may exist between fairness and utility, as well as between long-term and short-term fairness. The long-term fairness focuses on remedying past discrimination existed in the system, but has no constraint on the biases in the decision at each time step. The short-term fairness, on the other hand, cares about fairness in the decision making process at each time step, but pays no attention in correcting past discrimination in the population. One should combine long-term and short-term fairness to force the decision model to take into consideration both factors and to remove discrimination in the system gradually with time. Therefore, we similarly propose quantitative notions for short-term fairness and institution utility as follows.

\begin{definition}[Short-term Fairness] 
The short-term fairness of a decision model $h_{\theta}(\cdot)$ at time $t$ is measured by the causal effect transmitted through paths involved in time $t$, i.e., $P(\hat{Y}^t(s^+_{\pi^{t}},\theta)) - P(\hat{Y}^t(s^-_{\pi^{t}},\theta))$, where $\pi^{t} = \{S\rightarrow \tilde{\X}_{r} \rightarrow \hat{Y}^{t}, S \to \hat{Y}^t\}$ with redlining attributes $\tilde{\X}_{r}$, $s_{\pi}$ is the path-specific hard intervention and $\theta$ represents the soft intervention.
\end{definition}

\begin{definition}[Institution Utility] 
The institution utility of decision model $h_{\theta}(\cdot)$ is measured by the aggregate loss given by $\sum_{t=1}^{t*}\mathbb{E}[\mathcal{L}(Y^t, \hat{Y}^t)]$ where $\mathcal{L}(\cdot)$ is the loss function.
\end{definition}

\section{Learning Fair Decision Models}
After formulating related notions, we are ready to formulate the fair sequential decision making problem given a time-lagged causal graph. To ease the representation, in following discussions we consider the simplified causal graph shown in Figure \ref{fig:example2} where only relevant attributes with no redlining attributes exist. In this case, the long-term fairness is captured by paths from $S$ to $\hat{Y}^{t^{*}}$ through $\X^{1},\hat{Y}^{1},\cdots,\X^{t^{*}}$ as shown in red, and the short-term fairness is captured by the direct edge $S\rightarrow \hat{Y}^{t}$ at each time $t$ as shown in green. However, all our discussions can be applied to our general formulation that includes both relevant and irrelevant features.

The goal is to learn a functional mapping $h_{\theta}: (\mathbf{X}^t, S) \mapsto Y^t$ parameterized with $\theta$, i.e., $\hat{Y}^t = h_{\theta}(\mathbf{X}^t, S)$. Based on the discussions above, we formulate a constrained optimization problem which minimizes the loss while subject to long-term fairness and short-term fairness constrains simultaneously. The thresholds $\tau_l$ and $\tau_t$ control the strictness of constraints.

\begin{formulation}
\label{formualation1}
The problem of fair sequential decision making is formulated as the constrained optimization:
\begin{equation*}
\begin{split}
& \qquad \qquad \argmin_{\theta} \sum_{t=1}^{t^*} \mathbb{E} \left[ \mathcal{L}( Y^{t}, \hat{Y}^{t} ) \right] \\
\textrm{s.t.}& ~~ P\! \left( \hat{Y}^{t^{*}}(s^{+}_{\pi},\theta)\!=\!1 \right) - P\! \left( \hat{Y}^{t^*}(s^{-}_{\pi},\theta)\!=\!1 \right) \leq \tau_{l}, \\
& ~~ P\! \left( \hat{Y}^{t}(s^{+}_{\pi^{t}},\theta)\!=\!1 \right) - P\! \left( \hat{Y}^{t}(s^{-}_{\pi^{t}},\theta)\!=\!1 \right) \leq \tau_{t}, \\
& ~~ t=1,\cdots,t^*
\end{split}
\end{equation*}
where $\tau_l$ and $\tau_t$ are thresholds for long-term fairness and short-term fairness constraints, respectively.
\end{formulation}

\subsection{Formulating as Performative Risk Optimization}
Solving the optimization problem in Problem Formulation 1 is not trivial. According to the path-specific effect inference \cite{avin2005identifiability} and the definition of soft intervention \cite{correa2020calculus}, post-intervention probability $P(\hat{Y}^{t^{*}}(s^{+}_{\pi},\theta)=1)$ is given by

\begin{equation}\label{eq:inf}
\begin{split}
&     \sum_{\X^{1},Y^{1},\cdots,\X^{t^{*}}} \Big\{ P(\x^{1}|s^{+}) P_{\theta}(y^{1}|\x^{1},s^{-}) \cdots  \\
& \qquad  \cdots P(\x^{t^{*}}|\x^{t^{*}\!-\!1},y^{t^{*}\!-\!1}) P_{\theta}(Y^{t^{*}}\!=\!1|\x^{t^{*}},s^{-}) \Big\},
\end{split}
\end{equation}
where $P_{\theta}(y|\x,s^{-})$ is a probabilistic function determined by $h_{\theta}(\cdot)$. As a result, $P(\hat{Y}^{t^{*}}(s^{+}_{\pi},\theta)=1)$ is a complex nonlinear function of $\theta$, making Problem Formulation 1 difficult to solve.
In the following, we show how Problem Formulation 1 is converted to a performative risk optimization problem and then propose an optimization algorithm by leveraging repeated risk minimization.

Following the notation of convex optimization of classification, we denote by $\phi$ a convex surrogate function. Then, we can formulate the loss function as
\begin{equation*}
    \mathcal{L}( Y^{t}, \hat{Y}^{t} ) = \mathds{1}\left[ Y^{t}h_{\theta}(\X^{t},S)<0 \right] =  \phi\left( Y^{t}h_{\theta}(\X^{t},S) \right).
\end{equation*}

We can also apply the surrogate function to the fairness constraints. For any $t$,
we have
\begin{equation*}
\begin{split}
 & P\! \left( \hat{Y}^{t}(s^{+}_{\pi},\theta)\!=\!1 \right) 
 = \sum_{\X^{t}} P\! \left( \x^{t}(s^{+}_{\pi},\theta) \right) P_{\theta}(Y^{t}\!=\!1|\x,s^{-}).
\end{split}
\end{equation*}
Similar to \cite{hu2020fair}, we estimate $P_{\theta}(Y^{t}\!=\!1|\x,s^{-})$ by first treating it as $\mathds{1} \left[h_{\theta}\left(\x^{t},s^-\right)\geq 0 \right]$ and then replacing the indicator function by $\phi(\cdot)$:
\begin{equation*}
\begin{split}
 & P\! \left( \hat{Y}^{t}(s^{+}_{\pi},\theta)\!=\!1 \right) 
 = \sum_{\X^{t}} P\! \left( \x^{t}(s^{+}_{\pi},\theta) \right) \phi \left(-h_{\theta}\left(\x^{t},s^-\right)\right) \\
 & = \mathop{\mathbb{E}}_{ \scalebox{0.5}{$\X^{t}\sim P\! \left( \X^{t}(s^{+}_{\pi},\theta) \right)$} } \left[ \phi \left(-h_{\theta}\left(\X^{t},s^-\right)\right) \right].
\end{split}
\end{equation*}

Similarly, we have

\begin{equation*}
\begin{split}
 & -P\! \left( \hat{Y}^{t}(s^{-}_{\pi},\theta)\!=\!1 \right) = P\! \left( \hat{Y}^{t}(s^{-}_{\pi},\theta)\!=\!0 \right) -1 \\
 & = \mathop{\mathbb{E}}_{ \scalebox{0.5}{$\X^{t}\sim P\! \left( \X^{t}(s^{-}_{\pi},\theta) \right)$} } \left[ \phi \left(h_{\theta}\left(\X^{t},s^-\right)\right) \right] -1.
\end{split}
\end{equation*}

Then, we define utility loss $l_{u}(\theta)$, long-term fairness loss $l_{l}(\theta)$, and short-term fairness loss $l_{s}(\theta)$ as follows.

\begin{equation*}
    l_{u}(\theta) = \sum_{t=1}^{t^*} \mathop{\mathbb{E}}_{ \scalebox{0.5}{$S,\X^{t},Y^{t}\sim P(S,\X^{t},Y^{t})$} } \left[ \phi\left( Y^{t}h_{\theta}(\X^{t},S) \right) \right],
\end{equation*}

\begin{equation*}
\begin{split}
&    l_{l}(\theta) = \frac{1}{2} \left\{ \mathop{\mathbb{E}}_{ \scalebox{0.5}{$\X^{t^*}\sim P\! \left( \X^{t^*}(s^{+}_{\pi},\theta) \right)$} } \left[ \phi \left(-h_{\theta}\left(\X^{t^*},s^-\right)\right) \right] \right. \\
&  \left.  + \mathop{\mathbb{E}}_{ \scalebox{0.5}{$\X^{t^*}\sim P\! \left( \X^{t^*}(s^{-}_{\pi},\theta) \right)$} } \left[ \phi \left(h_{\theta}\left(\X^{t^*},s^-\right)\right) \right] - 1-\tau_{l} \right\} ,
\end{split}
\end{equation*}

\begin{equation*}
\begin{split}
&    l_{s}(\theta) = \frac{1}{t^*}\sum_{t=1}^{t^*} \left\{ \mathop{\mathbb{E}}_{ \scalebox{0.5}{$\X^{t}\sim P\! \left( \X^{t}(s^{-}_{\pi^{t}},\theta) \right)$} } \left[ \phi \left(-h_{\theta}\left(\X^{t},s^+\right)\right) \right] \right. \\
&   \left. + \mathop{\mathbb{E}}_{ \scalebox{0.5}{$\X^{t}\sim P\! \left( \X^{t}(s^{-}_{\pi^{t}},\theta) \right)$} } \left[ \phi \left(h_{\theta}\left(\X^{t},s^-\right)\right) \right] - 1 -\tau_{t}\right\}.
\end{split}
\end{equation*}

By adding the long-term and short-term fairness losses as regularization terms into the objective function, we obtain an unconstrained optimization problem as given in Problem Formulation 2. The general formulation of the performative risk optimization can be given by $\argmin_{\theta} \mathop{\mathbb{E}}\limits_{\Z \sim \mathcal{D}(\theta)} l(\Z; \theta)$ where $\Z$ represents the set of all attributes and outcome \cite{perdomo2020performative}. Thus, Problem Formulation 2 can be considered as a performative risk optimization problem as all terms in the objective function are represented as expectations of the loss function over the distributions that depend on the loss function parameters. Compare with Problem Formulation 1, Problem Formulation 2 relaxes the fairness constraints and certain amount of violations to the constraints are allowed. However, Problem Formulation 2 can be solve more efficiently by leveraging the repeated risk minimization technique as shown in the next subsection.

\begin{formulation}
\label{formulation2}
The problem of fair sequential decision making is reformulated as the performative risk optimization:
\begin{equation}
\label{loss}
    \argmin_{\theta} l(\theta) = \lambda_{u}l_{u}(\theta) + \lambda_{l}l_{l}(\theta) + \lambda_{s}l_{s}(\theta)
\end{equation}
where $\lambda_{u}$, $\lambda_{l}$ and $\lambda_{s}$ are weight parameters and satisfy $\lambda_{u}+\lambda_{l}+\lambda_{s}=1$.
\end{formulation}

\subsection{The Algorithm of Repeated Risk Minimization}
Repeated risk minimization (RRM) is an iterative algorithmic heuristic for solving the performative risk optimization problem. 
The procedure of the RRM is to start from an initial model and repeatedly find a model that minimizes the loss function on the distribution resulting from the previous model, which can symbolically represented as the update rule $\theta_{i+1} = \argmin_{\theta} \mathop{\mathbb{E}}\limits_{\Z \sim \mathcal{D}(\theta_i)} l(\Z; \theta)$ \cite{perdomo2020performative}. The RRM converges if the model that minimizes the loss remains unchanged from the previous model, i.e., $\theta_{i+1}=\theta_{i}$. 

To implement the RRM algorithm in our context with three different loss terms, we sample different distributions at each iteration. For computing $l_{u}(\theta)$, the data distribution does not change with the deployment of new models, and we always use the original dataset $\mathcal{D}$ to compute $l_{u}(\theta)$. For computing $l_{l}(\theta)$, the data distribution follows the post-intervention distribution $P(\X^{t^*}(s^{+}_{\pi},\theta))$ (resp. $P(\X^{t^*}(s^{-}_{\pi},\theta))$). Thus, we sample the data according to the inference formula that is similar to Eq.~\eqref{eq:inf} where a smooth probabilistic function $P_{\theta}(y|\x,s)$ is used. Specifically, we first sample $\X^{1}$ according to the distribution $P(\X^{1}|s^{+})$ (resp. $P(\X^{1}|s^{-})$), and sample the decision for each sample according to $P_{\theta}(Y^{1}|\x^{1},s^{-})$. Then, we sample $\X^{2}$ according to the distribution $P(\X^{2}|\X^{1},Y^{1})$ upon the samples obtained at the first time step. We repeat this process until time $t^{*}$ to obtain samples for $\X^{t^{*}}$ for computing $l_{l}(\theta)$. For computing $l_{s}(\theta)$, we similarly sample the distributions $P( \X^{t}(s^{+}_{\pi^{t}},\theta) )$ and $P( \X^{t}(s^{-}_{\pi^{t}},\theta) )$ for each time step $t$. The procedure of our algorithm starts from an initial model $h_{\theta_0}$ directly trained on $\mathcal{D}$, and repeatedly train the model on the re-sample data at each iteration, until the model converges to performative stability. The pseudocode of this procedure is described in Algorithm \ref{algo}.

\IncMargin{1em}
\begin{algorithm}[t]\small
\SetKwInOut{Input}{Input}
\SetKwInOut{Output}{Output}
\SetKwRepeat{Repeat}{repeat}{until}
\SetKw{Return}{return}
	\caption{Repeated Risk Minimization}
    \label{algo} 
    
	\Input{Dataset $\mathcal{D} = \{(S, \mathbf{X}^t, Y^t)\}_{t=1}^{l}$, time-lagged causal graph $\mathcal{G}$, convergence threshold $\delta$}
	\Output{The stable model $h_{\theta}$}
	\BlankLine 
	
	Train a classifier on $\mathcal{D}$ according to Eq. \eqref{loss} without the soft intervention to obtain the initial parameter $\theta_0$\;
	$i \leftarrow 0$\;
	\Repeat{$ \bigtriangleup < \delta$} {
	   
	   Sampled the post-intervention distributions $P\left( \X^{t^*}(s_{\pi}^{+},\theta_i) \right)$ and $P\left( \X^{t^*}(s_{\pi}^{-},\theta_i) \right)$\;
	   Sampled the post-intervention distributions $P\left( \X^{t}(s_{\pi}^{+},\theta_i) \right)$ and $P\left( \X^{t}(s_{\pi}^{-},\theta_i) \right)$ for each $t$\;
	   Minimize $l(\theta)$ according to Eq. \eqref{loss} to obtain $\theta_{i+1}$\;
	   $\bigtriangleup = \| \theta_{i+1} - \theta_{i} \|_2$\;
	   $i \rightarrow i+1$\;
	   
	}
    $\theta \leftarrow \theta_i$; \\
    \Return $h_{\theta}$;
 	 	  
\end{algorithm}
\DecMargin{1em}

\subsection{Convergence Analysis of RRM}
We now conduct performative stability analysis for our algorithm. 
The convergence of the RRM algorithm depends on the smoothness and convexity of the loss function, as well as the sensitivity of the distribution to the parameters \cite{perdomo2020performative}. Specifically, given a general RRM formulation $\theta_{i+1} = \argmin_{\theta} \mathop{\mathbb{E}}\limits_{\Z \sim \mathcal{D}(\theta_i)} l(\Z; \theta)$,
if loss function $l(\cdot)$ is $\beta$-jointly smooth and $\gamma$-strongly convex, and distribution $\mathcal{D}(\theta)$ is $\varepsilon$-sensitive, then the RRM converges to a stable point if $\varepsilon < \frac{\beta}{\gamma}$.
We similarly analyze these factors for our problem and then give the theoretical convergence result.

\begin{lemma}\label{thm:l1}
If the surrogated loss function $(\phi\circ h)(\cdot)$ is $\gamma$-strongly convex, then $l(\cdot)$ is $\gamma$-strongly convex.
\end{lemma}

Lemma \ref{thm:l1} can be directly proven according to the sum rule of the gradient. 

Next, we study the sensitivity of the distributions. Consider the distribution $P(\X^{t}(s_{\pi},\theta))$ for any $t$. Its sensitivity to $\theta$ depends on to what extend the decisions will impact the attributes via the feedback loop. By assuming that the change of the distribution over the attributes in respond to the change of $\theta$ is bounded by a constant, we present following lemma. Please refer to the appendix for detailed proof.

\begin{definition}
For any $t$, attributes $\X^{t+1}$ are $c$-sensitive if 
\begin{equation*}
\begin{split}
        &\| \sum_{Y^{t}}\nabla_{\theta} P_{\theta}(y^{t}|\x^{t},s)P(\x^{t+1}|\x^{t},y^{t}) \| \\
        &\leq c \sum_{Y^{t}} P(\x^{t+1}|\x^{t},y^{t}).
\end{split}
\end{equation*}
\end{definition}

\begin{lemma}\label{thm:l2}
For any $t$, suppose that $\X^{t+1}$ are $c$-sensitive, then distribution $P(\X^{t}(s_{\pi},\theta))$ is $\varepsilon$-sensitive with 
    $\varepsilon \leq 2mc(t-1)$,
where $m$ is the maximum ground distance between two values of $\X^{t}$.
\end{lemma}

After introducing the above two lemmas, we now present our main theoretical result.

\begin{theorem}\label{thm:t1}
Suppose that surrogated loss function $(\phi\circ h) (\cdot)$ is $\beta$-jointly smooth and $\gamma$-strongly convex, and suppose that $\X^{t+1}$ are $c$-sensitive for any $t$, then the repeated risk minimization converges to a stable point at a linear rate, 
if $2mc(t^{*}-1) < \frac{\beta}{\gamma}$.
\end{theorem}

The proof of Theorem \ref{thm:t1} is based on Theorem 3.5 in \cite{perdomo2020performative}. Please refer to the appendix for details. In practice, this theoretical criterion of convergence may be difficult to meet. However, our experimental results show that our algorithm can converge under reasonable conditions.

\section{Experiments}
We conduct experiments on both synthetic and semi-synthetic temporal datasets to evaluate the proposed algorithm. We show that our algorithm is effective in achieving both long-term and short-term fairness, while previous fair algorithms that do not consider the dynamics in sequential decision making actually do not mitigate or even exacerbate the short-term or long-term fairness. We consider three baselines in the experiments which treat the whole temporal dataset as a static dataset and train the decision model on it. Fairness constraints are added following the technique proposed in \cite{wu2019convexity}.

\begin{itemize}
    \item \textbf{Logistic Regression (LR)}: An unconstrained logistic regression model which takes user features and labels from all time steps as inputs and outputs.
    
    \item \textbf{Fair Model with Demographic Parity (FMDP)}: On the basis of the logistic regression model, fairness constraint is added to achieve demographic parity. 
    \item \textbf{Fair Model with Equal Opportunity (FMEO)}: On the basis of the logistic regression model, fairness constraint is added to achieve equal opportunity. 
\end{itemize}

\subsection{Datasets}
{\bf\noindent Synthetic Data.}
We simulate a process of bank loans following the time-lagged causal graph depicted in Figure \ref{fig:example2}, where $S$ is the protected attribute like race, $\mathbf{X}^t$ represents the financial status of applicants, and $Y^t$ represents the decisions about whether or not to grant loans. At $t=1$, we generate samples where both values of $S$ are sampled with the equal probability, and the values of $\X^{1}$ are sampled using two different Gaussian distributions according to the value of $S$.
Then at each time $t$, we sample  predicted decisions $\hat{Y}^{t}$ and the values of $\X^{t+1}$ as follows. Consider a ground-truth decision model $h_{\theta^{*}}(\cdot)$ for deciding the probability of whether an individual would default on a loan, given by $\sigma(h_{\theta^{*}}(\cdot))$ where $\sigma(\cdot)$ is the sigmoid function. Then, we sample the predicted decision $\hat{Y}^{t}$ (as well as the actual repayment $Y^{t}$ which is sampled separately) from $\sigma(h_{\theta^{*}}(\cdot))$ as:
\begin{equation*}
    \begin{split}
    & P(\hat{Y}^t) = \sigma(h_{\theta^*}(\mathbf{X}^t, S)), \thinspace \hat{Y^t} \sim 2\cdot \text{Bernoulli}(P(\hat{Y}^t))-1.
    \end{split}
\end{equation*}
Then, $\X^{t+1}$ is generated according to the update rule below:
\begin{equation}
    \label{eq:changes}
    \mathbf{X}^{t+1} =  \left\{
    \begin{array}{lcl}
    \mathbf{X}^t - \epsilon \cdot \theta^t + b & & \hat{Y}^t=1, Y^t=-1 \\
    \mathbf{X}^t + \epsilon \cdot \theta^t + b & & \hat{Y}^t=1, Y^t=1 \\
    \mathbf{X}^t + b & & \hat{Y}^t = -1
    \end{array}
    \right.
\end{equation}
where $\epsilon$ is a parameter that controls the sensitivity of the update to the predicted decisions, and $b = S \cdot b_1 + (1 - S) \cdot b_0$ is a small base increment at each time step.
In the simulation process, we generate a 5-step synthetic dataset with 5000 samples where parameters are set as $\epsilon = 0.5$, $b_0 = 0.2$, $b_1 = 1.0$.

{\bf\noindent Semi-synthetic Data.} We use the Taiwan credit card dataset \cite{yeh2009comparisons} as the initial data at $t=1$. To form a balanced dataset, we extract 3000 samples and choose two features PAY\_AMT1 and PAY\_AMT2 that are appropriate in fitting into our update rule. Then, we generate a 4-step dataset using the same update rule as shown above.

\subsection{Training and Evaluation}
We conduct the training process following the RRM algorithm. At each iteration, we sample the data according to the current decision model and the causal graph. Similar to the data generation process, predicted decisions are sampled according to the probability given by $\sigma(h_{\theta}(\cdot))$, and the feature values are sampled according to Eq.~\eqref{eq:changes}. In our experiments, we assume that the true update rule is known in order to remove errors introduced by causal graph construction. In practice, the causal graph learned from data may introduce additional errors.

We then design an evaluation process which simulates the real model deployment procedure and feedback loops. At each time step $t$, we use the trained decision model $h_{\theta}(\cdot)$ to make decisions $\hat{Y}^{t}$, and use the ground-truth model $h_{\theta^{*}}(\cdot)$ to determine the repayment $Y^{t}$. The accuracy is measured by comparing $\hat{Y}^{t}$ and $Y^{t}$, the long-term fairness is measured based on the distribution of $\hat{Y}^{t^{*}}$ in the evaluation, and the short-term fairness is measured based on the distribution of $\hat{Y}^{t}$ at different time steps according to proposed definitions.

\begin{table}[!ht]\small
	\centering
	\begin{tabular}{c|c|c|c|c|c|c}
        \Xhline{1pt}
        \multirow{2}{*}{Alg.} & \multirow{2}{*}{Metric} &
        \multicolumn{5}{c}{Time steps} \\
        \cline{3-7}
        & & $t\!=\!1$ & $t\!=\!2$ & $t\!=\!3$ & $t\!=\!4$ & $t\!=\!5$ \\
        \Xhline{0.8pt}
        \multirow{3}{*}{RL} & Acc & 0.912 & 0.894 & 0.917 & 0.921 & 0.917\\
        \cline{2-7}
        & Short & 0.152 & 0.160 & 0.166 & 0.164 & 0.174 \\
        \cline{2-7}
        & Long & 0.058 & 0.117 & 0.173 & 0.246 & 0.340 \\
        \hline
        \multirow{3}{*}{FMDP} & Acc & 0.735  & 0.706 & 0.704 & 0.708 & 0.725 \\
        \cline{2-7}
        & Short & 0.212 & 0.216 & 0.224 & 0.220 & 0.232 \\
        \cline{2-7}
        & Long  & 0.180 & 0.306 & 0.376 & 0.431 & 0.481 \\
        \hline
        \multirow{3}{*}{FMEO} & Acc & 0.829  & 0.790 & 0.795 & 0.800 & 0.814\\
        \cline{2-7}
        & Short & 0.010 & 0.010 & 0.010 & 0.014 & 0.020\\
        \cline{2-7}
        & Long  & 0.080 & 0.122 & 0.190 & 0.276 & 0.352\\
        \Xhline{0.7pt}
        \multirow{3}{*}{Ours} & Acc & 0.801  & 0.754 & 0.729 & 0.707 & 0.692\\
        \cline{2-7}
        & Short & 0.012 & 0.008 & 0.012 & 0.008 & 0.002\\
        \cline{2-7}
        & Long  & 0.040 & 0.024 & 0.020 & 0.012 & 0.002\\
        \Xhline{1pt}
	\end{tabular}
	\caption{Accuracy, short-term and long-term fairness of different algorithms on the synthetic dataset.}
	\label{tab:exp_syn}
\end{table} 

\begin{table}[!ht]\small
	\centering
	\begin{tabular}{c|c|c|c|c|c}
        \Xhline{1pt}
        \multirow{2}{*}{Alg.} & \multirow{2}{*}{Metric} &
        \multicolumn{4}{c}{Time steps} \\
        \cline{3-6}
        & & $t\!=\!1$ & $t\!=\!2$ & $t\!=\!3$ & $t\!=\!4$ \\
        \Xhline{0.8pt}
        \multirow{3}{*}{RL} & Acc & 0.828 & 0.826 & 0.841 & 0.816 \\
        \cline{2-6}
        & Short & 0.015 & 0.018 & 0.021 & 0.012 \\
        \cline{2-6}
        & Long & 0.038 & 0.088 & 0.243 & 0.433   \\
        \hline
        \multirow{3}{*}{FMDP} & Acc & 0.830  & 0.843 & 0.846 & 0.841  \\
        \cline{2-6}
        & Short & 0.063 & 0.066 & 0.075 & 0.069  \\
        \cline{2-6}
        & Long  & 0.038 & 0.076 & 0.223 & 0.397  \\
        \hline
        \multirow{3}{*}{FMEO} & Acc & 0.824  & 0.830 & 0.830 & 0.813  \\
        \cline{2-6}
        & Short & 0.072 & 0.075 & 0.087 & 0.078  \\
        \cline{2-6}
        & Long  & 0.006 & 0.045 & 0.156 & 0.295  \\
        \Xhline{0.7pt}
        \multirow{3}{*}{Ours} & Acc & 0.648  & 0.648 & 0.680 & 0.687  \\
        \cline{2-6}
        & Short & 0.006 & 0.006 & 0.003 & 0.006 \\
        \cline{2-6}
        & Long  & 0.064 & 0.043 & 0.016 & 0.003  \\
        \Xhline{1pt}
	\end{tabular}
	\caption{Accuracy, short-term and long-term fairness of different algorithms on the semi-synthetic dataset.}
	\label{tab:exp_semi}
\end{table} 

{\bf\noindent Implementation Details.\footnote{The code and hyperparameter settings are available online: \url{https://github.com/yaoweihu/Achieving-Long-term-Fairness}.}}
For baselines FMDP and FMEO, they are formulated as constrained optimization forms which are directly solved by the CVXPY package \cite{diamond2016cvxpy}. For our algorithm, we use the logistic loss function for the surrogate function $\phi$ and the linear model for the decision model. All algorithms use the $l_2$-regularization which can equip the logistic loss function with strong convexity. In our algorithm, ReLU activation function is adopted to ensure that the fairness constraints are always non-negative, and we adopt PyTorch \cite{NEURIPS2019_9015} to implement optimization with Adam optimizer.

\subsection{Results}
The results of the accuracy and fairness of the baselines and our algorithm on the synthetic dataset are shown in Table \ref{tab:exp_syn}. As can be seen, our algorithm achieves the short-term fairness at all time steps. More importantly, the long-term fairness is improved with time and approaches zero at $t=5$. For other baselines, there is a clear trend that the long-term fairness continuously accumulates with time.
This demonstrates that static fairness notions may harm fairness in the long run. The short-term fairness remains stable with time as it shows the bias in the model that is related to the protected attribute. The experiments on the semi-synthetic dataset produce similar results as shown in Table \ref{tab:exp_semi}. We also observe a trade-off between accuracy and fairness meaning that some accuracy needs to be sacrificed in order to achieve fairness.

\begin{figure}[t]
    \centering
	\includegraphics[width=7cm]{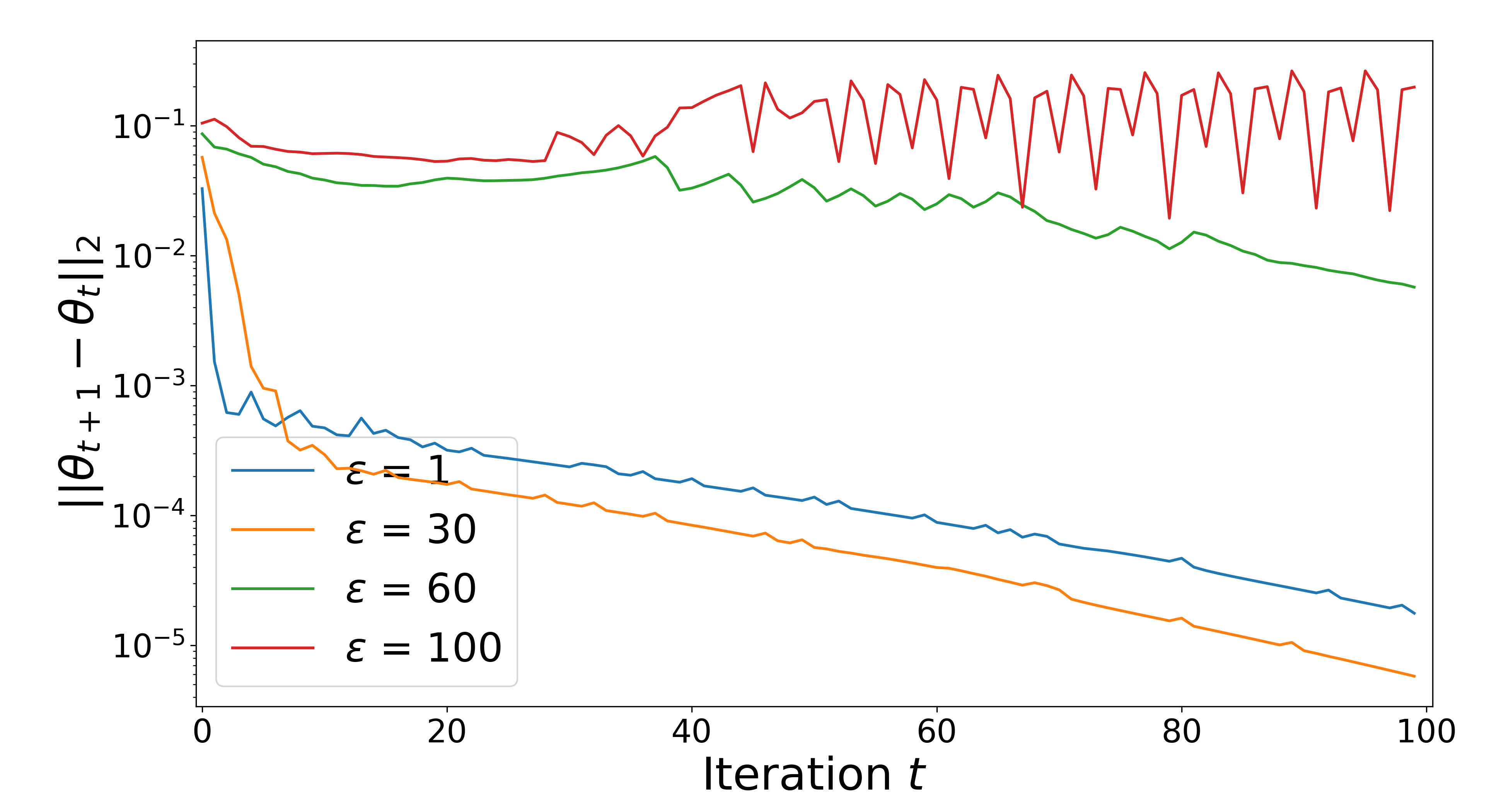}
	\caption{The convergence results for different values of $\epsilon$ on the synthetic dataset.}
	\label{fig:convergence}
\end{figure}

We also plot in Figure \ref{fig:convergence} the convergence results of our algorithms for different $\epsilon$ values. As mentioned earlier, the value of $\epsilon$ controls the sensitivity of $\X^{t+1}$ to the update of $\theta$. Figure \ref{fig:convergence} shows that our algorithm converges when the value of $\epsilon$ is reasonably small, which is consistent with the results in \cite{perdomo2020performative}. We observe similar results on the semi-synthetic dataset.

\section{Conclusions and Future Work}
We proposed a framework to achieve long-term fairness in sequential decision making. The decision-making process was modeled by a time-lagged causal graph, in which the hard intervention was performed on the protected attribute and soft interventions were performed on the decisions. We measured both long-term and short-term fairness as path-specific effects. The problem of fair sequential decision making was formulated as 
a performative risk optimization problem, and repeated risk minimization is adopted to train the model on the datasets sampled from post-intervention distributions. The convergence of the proposed algorithm is analyzed theoretically. Finally, we verify the effectiveness of the proposed framework and algorithm by comparing it with the baselines on two synthetic datasets.

Path-specific effects may be unidentifiable from the observational data if the ``kite structure'' presents in the causal graph \cite{avin2005identifiability}. The long-term fairness loss term $\l_{l}(\theta)$ cannot be accurately estimated if $P( \X^{t^*}(s^{+}_{\pi},\theta) )$ is unidentifiable if the paths in $\pi$ form a ``kite structure''. We will adopt the bounding technique proposed in \cite{zhang2018causal} for unidentifiable path-specific quantify, compute the lower and upper bounds of $P( \x^{t^*}(s^{+}_{\pi},\theta) )$ for each $\x$, and then obtain the upper bound of $\l_{l}(\theta)$. We leave this to our future work.

\newpage

\section*{Ethics Statement}
Our research could benefit decision makers for achieving long-term fairness and balancing the trade-off between fairness and accuracy.
The proposed method relaxes the constrained optimization problem (Problem Formulation 1) to an unconstrained optimization problem (Problem Formulation 2). There may be gaps between the two problems, i.e., the optimal solution to the unconstrained optimization problem may not be the optimal solution to the original constrained one. This may potentially result in solutions that achieve compromised fairness which is lower than user requirements.

\section*{Acknowledgments}
This work was supported in part by NSF 1910284, 1920920, and 1946391.

\bibliography{aaai22}

\end{document}